\documentclass{svproc}
\usepackage{url}

\usepackage{graphicx}

\begin{document}
	\mainmatter              
	\title{Towards a practical lip-to-speech conversion system using deep neural networks and mobile application frontend}
	\titlerunning{Lip-to-speech and mobile app}  
	%
	\author{Frigyes Viktor Arthur\inst{1} \and Tam\'{a}s G\'{a}bor Csap\'{o}\inst{1,2}}
	\authorrunning{Arthur \& Csap\'{o}} 
	%
	\tocauthor{Frigyes Viktor Arthur and Tam\'{a}s G\'{a}bor Csap\'{o}}
	\institute{Department of Telecommunications and Media Informatics,\\ Budapest University of Technology and Economics, Budapest, Hungary
		\and
		MTA-ELTE ,,Lend\"{u}ület'' Lingual Articulation Research Group, Budapest, Hungary\\
		\email{hello@victorarthur.com, csapot@tmit.bme.hu}}

	\maketitle              

	\begin{abstract}
		Articulatory-to-acoustic (forward) mapping is a technique to predict speech using various articulatory acquisition techniques as input (e.g. ultrasound tongue imaging, MRI, lip video). The advantage of lip video is that it is easily available and affordable: most modern smartphones have a front camera. There are already a few solutions for lip-to-speech synthesis, but they mostly concentrate on offline training and inference. In this paper, we propose a system built from a backend for deep neural network training and inference and a fronted as a form of a mobile application. Our initial evaluation shows that the scenario is feasible: a top-5 classification accuracy of 74\% is combined with feedback from the mobile application user, making sure that the speaking impaired might be able to communicate with this solution.
	\end{abstract}
	\section{Introduction}

	Speech sounds result from a coordinated movement of articulation organs (vocal cords, tongue, lips, etc.). The relationship between articulation and the resulting speech signal has been studied by machine learning tools. The results of the articulatory-to-acoustic (forward) mapping (AAM) contribute to the development of 'Silent Speech Interface' systems (SSI~\cite{Denby2010,Gonzalez-Lopez2020}). The essence of SSI is recording the articulation organs while the user of the device actually does not make a sound, but yet the machine system can synthesise speech based on the movement of the organs. In the long run, this potential application can contribute to the creation of a communication tool for speech-impaired people (e.g. those who lost voice after laryngectomy). Voice assistants are getting popular lately, but they are still not so widely used. One of the reasons is privacy concerns; some people do not feel comfortable if they have to speak loud, having others around. In this case, automatic lip-reading can be a solution.
	The relevance of this work is that such research about articulatory-to-acoustic mapping can also contribute to brain-computer interfaces (BCI)~\cite{Selim2020} and direct brain-to-speech conversion~\cite{Krishna2020}.

	\subsection{Lip-to-speech conversion}

	For the articulatory-to-acoustic conversion task, typically electromagnetic articulography~\cite{Wang2012a}, ultrasound tongue imaging~\cite{Csapo2017c,Csapo2020c}, permanent magnetic articulography~\cite{Gonzalez2017a}, surface electromyography~\cite{Janke2017}, magnetic resonance imaging~\cite{Csapo2020a} or video of the lip movements~\cite{LeCornu2015,Ephrat2017,Akbari2018,Racz2020,Michelsanti2020,Wand2016,Sun2018,Wand2020} are used. Lip-to-speech synthesis can be solved in two different ways: 1) direct approach, meaning that speech is generated without an intermediate step from the input signal~\cite{LeCornu2015,Ephrat2017,Akbari2018,Racz2020,Michelsanti2020}; and 2) indirect approach, meaning that lip-to-text recognition is followed by text-to-speech synthesis~\cite{Wand2016,Sun2018,Wand2020}. The direct approach has the advantage that potentially it can be faster as there are no intermediate steps in the processing. On the other hand, during the indirect approach, knowledge about audio, visual and face recognition can be used. Wand and his colleagues tested multiple approaches for lipreading and audiovisual speech recognition~\cite{Wand2016,Wand2020}. Recently, Lip-Interact was proposed, which is based on silent speech recognition and allows silent lip movement to be used for interactions with the smartphone~\cite{Sun2018}.

	\subsection{Deep learning in visual and audio recognition tasks}

	Recently, deep neural networks have demonstrated accuracy better than or equivalent to human performance in several different visual or audio recognition tasks, such as object detection~\cite{Ren2015}, image classification~\cite{Krizhevsky2012}, edge (contour) detection~\cite{Xie2015}, and speech technologies~\cite{Omar2020}. Most of the above lip-to-speech system also apply deep learning. For example, Le Cornu and Milner compare GMMs, DNNs and apply image preprocessing on the face images~\cite{LeCornu2015}. Ephrat and Peleg use convolutional neural networks for this task~\cite{Ephrat2017}. In~\cite{Akbari2018,Wand2016} and also in our earlier work~\cite{Racz2020}, both recurrent networks and CNNs are used. Most recently, architectures for modality fusion were also tested~\cite{Wand2020}.

	\subsection{Contributions of the current paper}

	Although there are several available solutions for lip-to-speech or lip-to-text conversion, they mostly focus on offline training and inference. In the current paper, we propose a practical system consisting of a backend for deep neural network training and real-time inference and using a mobile application frontend. To achieve the goal, we focused on CNN and LSTM architectures of deep neural networks and experimented with various parameters.


	\section{Methods}

	\begin{figure}
		\centering
		\includegraphics[width=\textwidth, keepaspectratio]{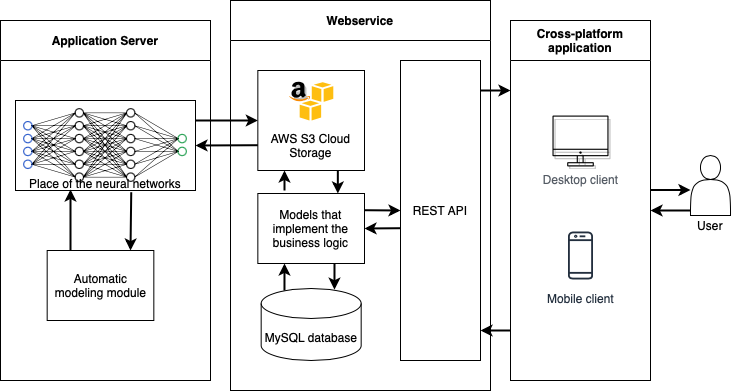}
		\caption{Structure of the system.}
		\label{fig:soft_struct}
	\end{figure}

	\subsection{Main components of the system}

	The proposed prototype system consists of three main components and their subsystems, as can be seen in Fig.~\ref{fig:soft_struct} and is detailed below.


	\subsubsection{Application server}

	The application server is storing the data related to machine learning and also the deep neural network experiments (both training and inference) are run in this component. After the training of the neural nets, the trained models and their weights are also stored here. During inference, the network is called with the trained weights and the new input data, and the resulting text is given to the web service in order to be sent to the client application.

	\subsubsection{Webservice}

	Between the applications server and mobile application, there is a web service, having the following tasks:
	user authentication
	data recording
	handling of function calls in the local database
	storing the training data within the cloud-based storage
	For our evaluations, the data was stored at AWS S3, the web service was hosted at DigitalOcean, and LetsEncrypt issued the SSL certificate.

	\subsubsection{Cross-platform mobile application}

	\begin{figure}
		\centering

		\includegraphics[width=0.35\textwidth, keepaspectratio]{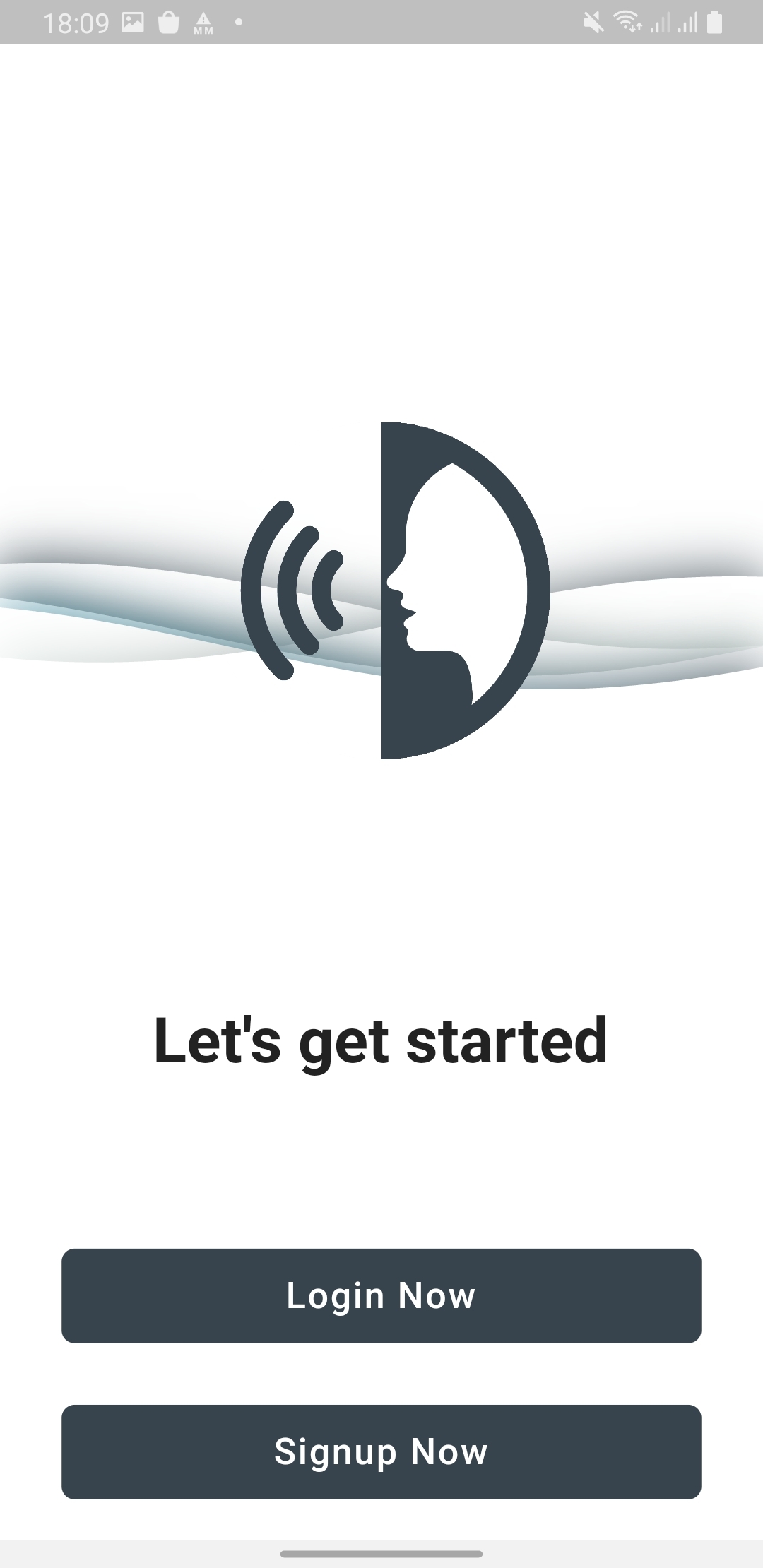}\qquad
		\includegraphics[width=0.35\textwidth, keepaspectratio]{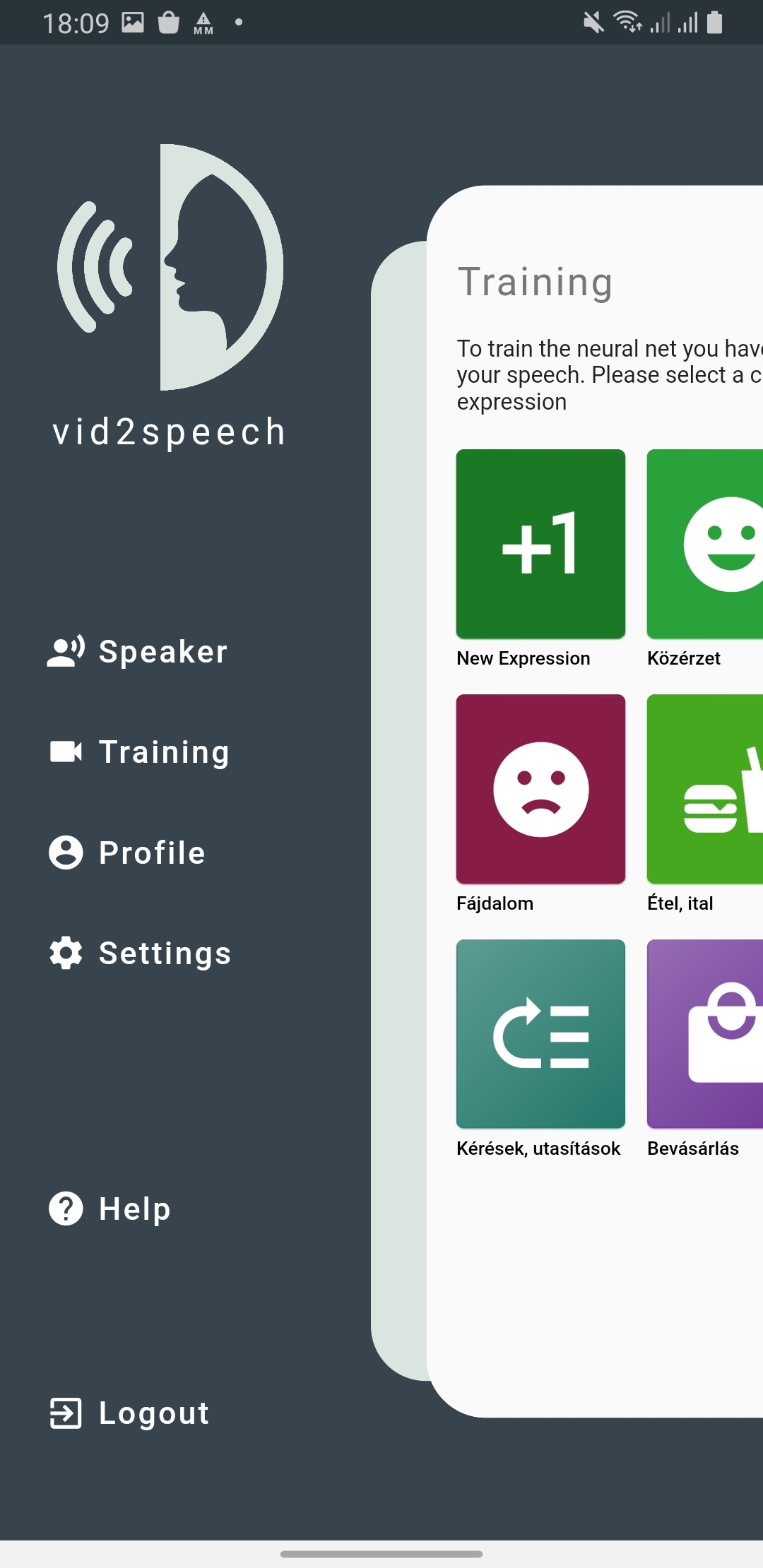}

		\includegraphics[width=0.35\textwidth, keepaspectratio]{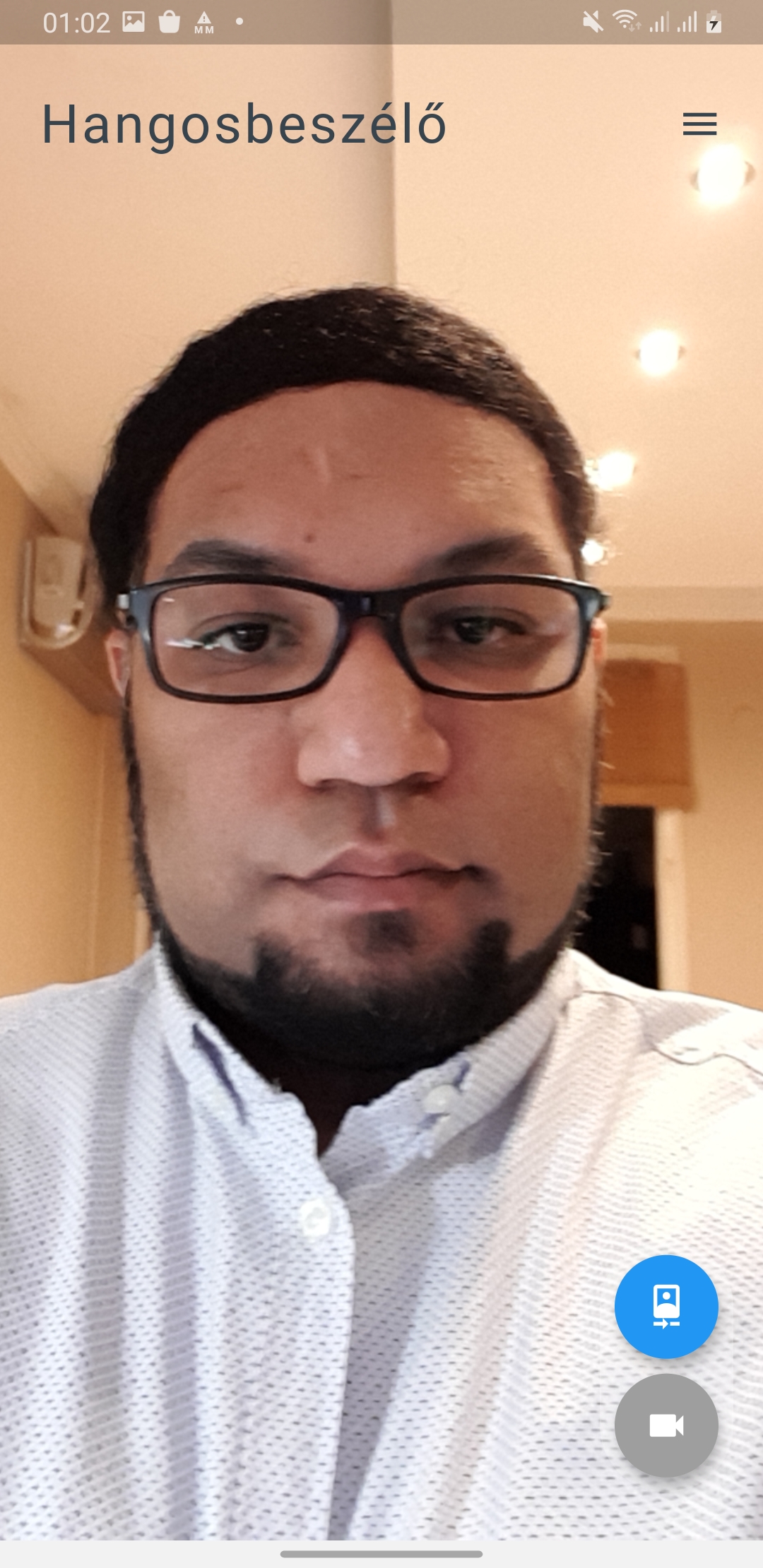}\qquad
		\includegraphics[width=0.35\textwidth, keepaspectratio]{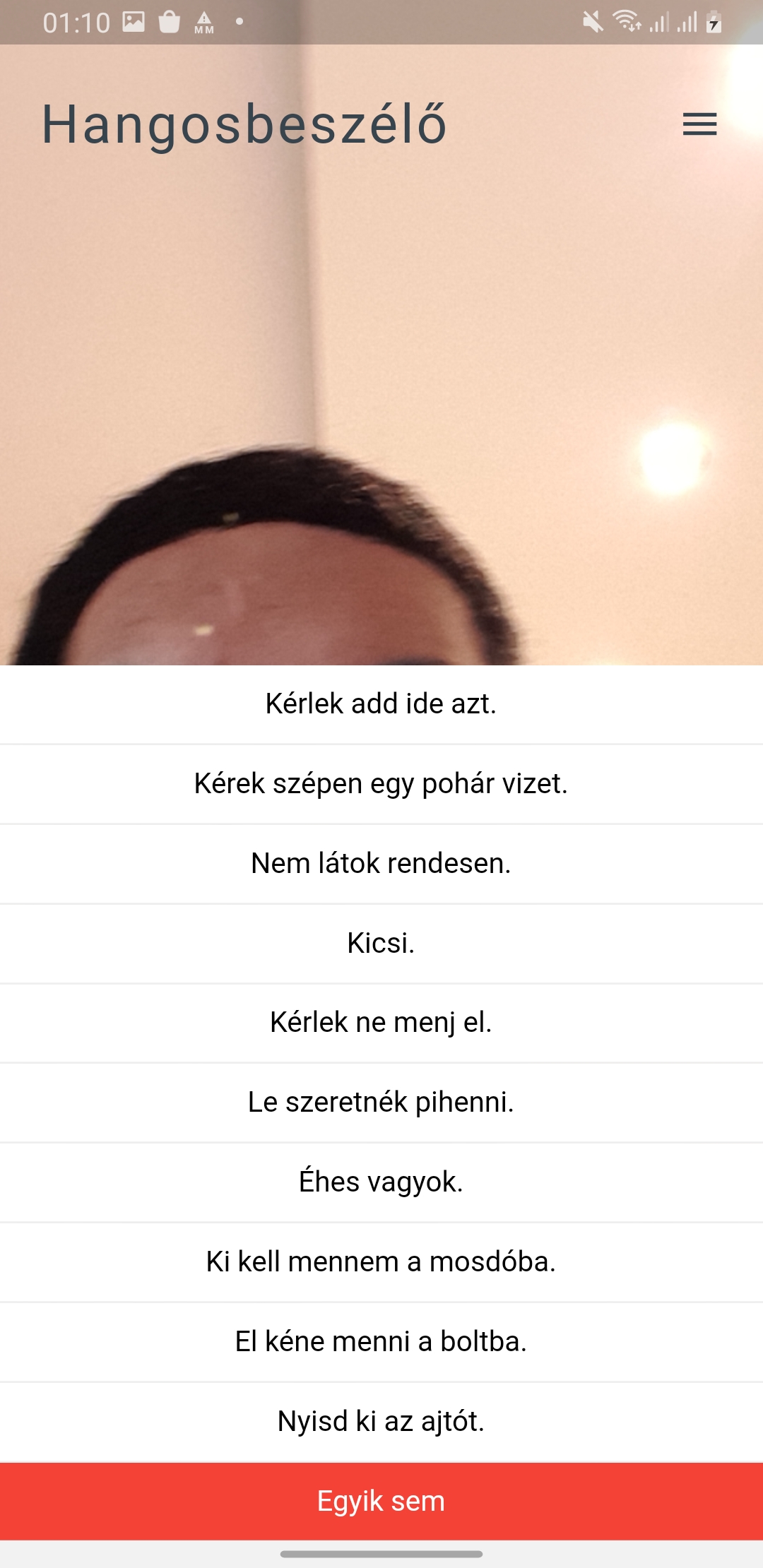}

		\caption{Screen captures from the mobile application (English and Hungarian versions). Top left: login screen. Top right: navigation menu. Bottom left: recording silent lip movement. Bottom right: showing the result of lip-to-text conversion (multiple results in a list).}
		\label{fig:mobile_app}

	\end{figure}

	The user can interact with the system through the cross-platform application, which was developed using the Flutter framework; and is therefore platform-independent and can be used on various mobile clients or desktop browsers. We tested this application on an emulator having Android 9.0 and on a Samsung Galaxy Note 9 device having Android 10.0 system during our evaluations. Fig.~\ref{fig:mobile_app} shows several sample screenshots from the English and Hungarian version of the mobile application. It has screens for registration, login, recording training data, uploading them to the server. Besides, in the 'loud speaking' mode, test video can be recorded within the application, after which the video is sent to the application server through the web service. The inference is run on the server, the result of which (the recognized text) is sent back to the mobile client. This text is sent to the system text-to-speech system within the mobile application -- this way, the application can be used for communication purposes.

	\subsection{Video recording within the mobile client}

	Both the training data and the video during the inference are recorded in the mobile client.

	For the textual content of the training data, we selected 88 Hungarian expressions that were proposed earlier for the StrokeAid application, being a communication tool for speaking impaired people (\url{https://play.google.com/store/apps/details?id=com.onlab.monddki}). Within the 'training' mode, we ask the users to record five repetitions of these sentences. The video of the face is recorded at 720x1280 pixels resolution and 25 fps using the front camera of the smartphone, and after recording each sentence, the data is sent to the application server through the web service. In our evaluation, we tested the training mode with a male speaker (being the first author of the paper), who read the 440 expressions.

	\subsection{Processing video data}

\begin{figure}
		\centering

		\includegraphics[height=0.20\textwidth, keepaspectratio]{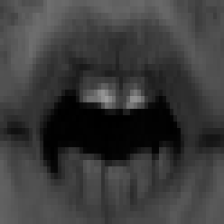}\hspace{5mm}	\includegraphics[height=0.20\textwidth, keepaspectratio]{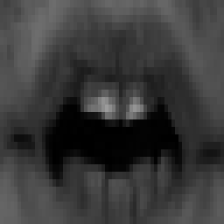}\hspace{5mm}	\includegraphics[height=0.20\textwidth, keepaspectratio]{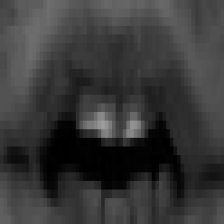}\hspace{5mm}	\includegraphics[height=0.20\textwidth, keepaspectratio]{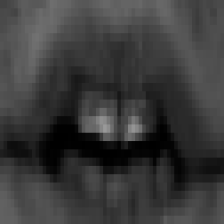}\hspace{5mm}
		\caption{Samples for lip data cut from the videos, which are the input of the DNNs.}

		\label{vgg16_minta}
	\end{figure}

	After the training data is collected in the mobile client, the remaining processing of the videos is done on the application server. For detecting the anchor points on the face, we tested three solutions: 2d106det MobileNet~\cite{deng2018menpo},  Google Firebase ML Kit FireVision \cite{noauthor_face_nodate}, and the 'shape\_predictor\_68\_face\_landmarks' model of DLib~\cite{noauthor_ibug_nodate}. After running speed and compatibility tests, we have selected DLib for our purposes. It returns 68 anchor points, based on which the mouth's region was cut out from the videos. Next, this region was resized to 299x299 pixels for the input of the neural network. A few sample lip images are shown in Fig.~\ref{vgg16_minta}.

	\subsection{DNN training within the application server}

\begin{figure}
		\centering
		\includegraphics[width=\textwidth, keepaspectratio]{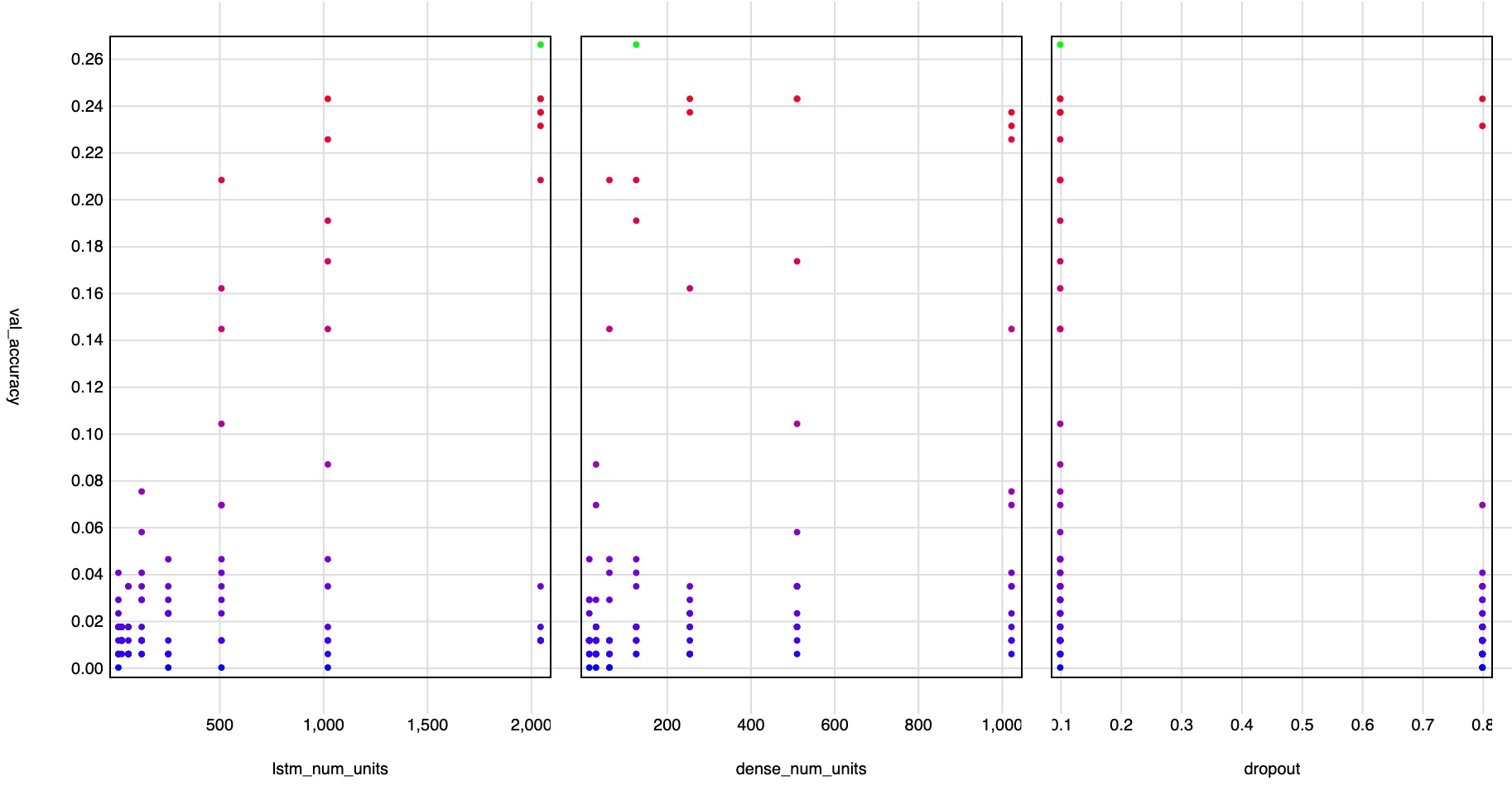}\hspace{5mm}
		\caption{Top-1 validation accuracy as a function of the hyperparameters.}
		\label{fig:hyperopt}
	\end{figure}

	The recorded data of 440 videos were separated into 60\% training and 40\% validation. We were using convolutional neural networks for feature extraction, whereas, for the final classification, recurrent networks were used. We applied automatic hyperparameter optimization, and the results are shown in Fig.~\ref{fig:hyperopt}. The optimal network structure is the following: InceptoonV3 for the feature extraction, followed by a single LSTM layer having 2048 neurons, a DropOut of 10\%, and a fully connected layer with 128 neurons at the end. During training, early stopping was applied with patience of 10 epochs. We trained the network in classification mode, using categorical cross-entropy cost function and ADAM optimizer (learning rate: $10^{-5}$).

	\section{Results and discussion}

	For testing purposes, new videos were recorded (a single utterance for each of the 88 Hungarian sentences), ensuring that the training and validation data are not used here.

\begin{figure}
		\centering
		\includegraphics[width=\textwidth, keepaspectratio]{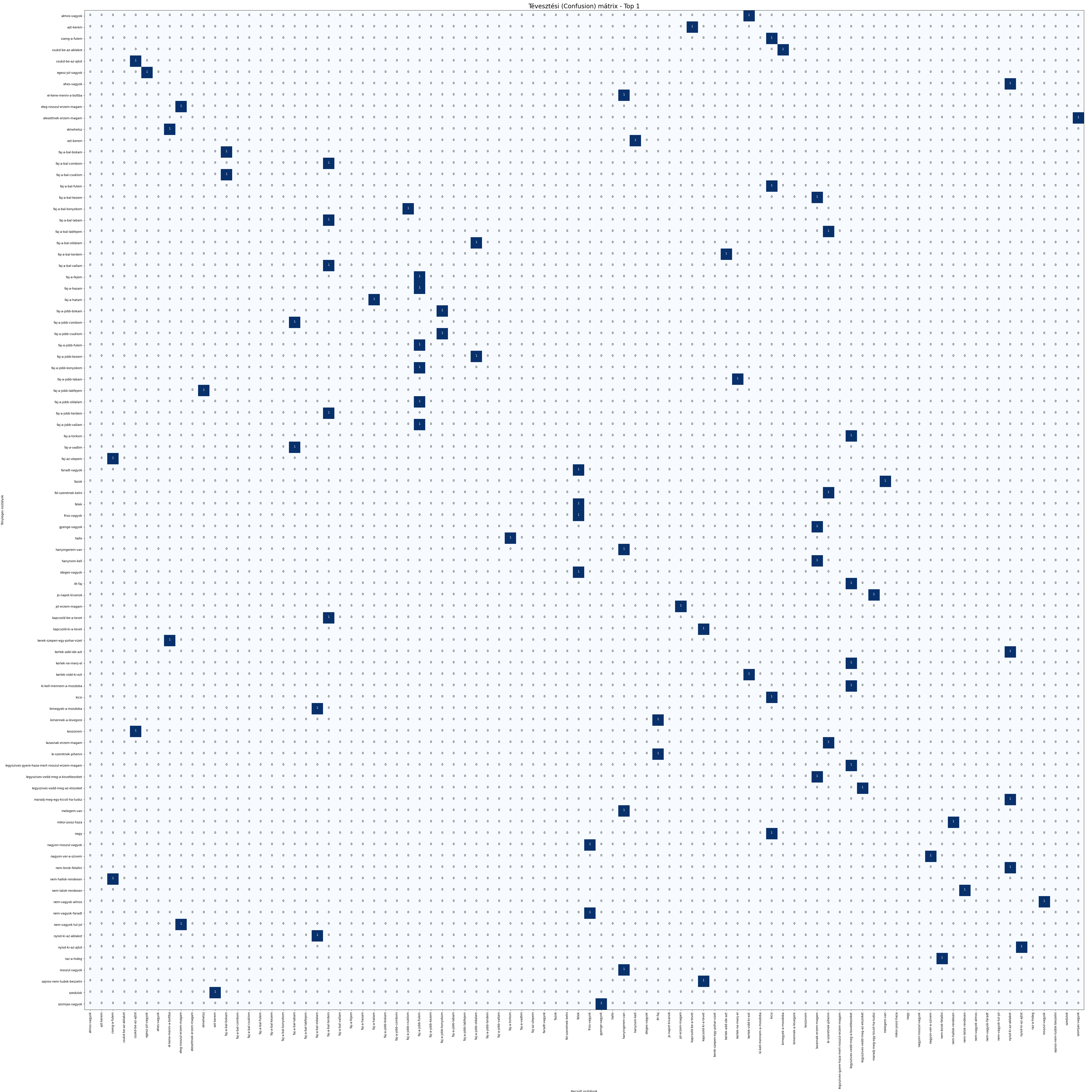}
		\caption{Top-1 confusion matrix. (full image at: \protect\url{https://github.com/victorarthur/vid2speech_images})}
		\label{fig:results_top1}
	\end{figure}

\begin{figure}
		\centering
		\includegraphics[width=\textwidth, keepaspectratio]{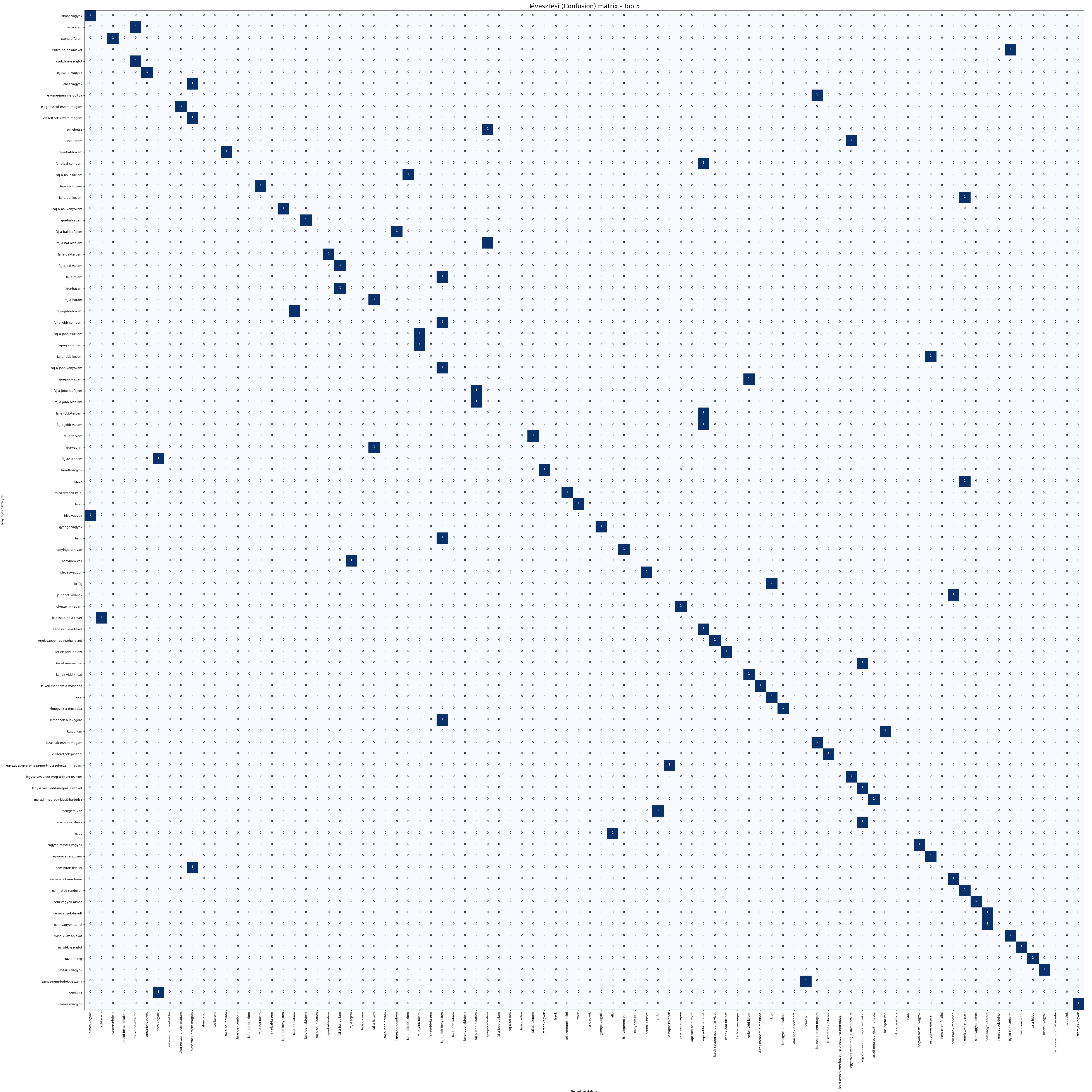}
		\caption{Top-5 confusion matrix. (full image at: \protect\url{https://github.com/victorarthur/vid2speech_images})}
		\label{fig:results_top5}
	\end{figure}

	As the results of the classification, the top-1 confusion matrix can be seen in Fig.~\ref{fig:results_top1}, while Fig.~\ref{fig:results_top5} shows the top-5 accuracies (full images at: \url{https://github.com/victorarthur/vid2speech_images}). Altogether, the final model achieved 53\% top-1 and 74\% of top-5 accuracy. We can compare this with human lipreading performance, which is about 30\%~\cite{altieri_normative_2011}. In the confusion matrices, the optimal case would be when many of the results are in the diagonal. In the case of top-1 accuracy (Fig.~\ref{fig:results_top1}), this was not achieved: in many cases; the silent lip videos were misrecognized. On the other hand, the top-5 classification matrix (Fig.~\ref{fig:results_top5}) contains more elements around the diagonal, showing that the network found the text that was uttered by the subject with acceptable performance.

	In the practical system, after the inference is run on the application server, the top recognition results (ordered based on their probability) are sent back to the mobile client, and the user can choose which was the uttered sentence (see Fig.~\ref{fig:mobile_app}, bottom right), before it is sent to the text-to-speech module of the system. This step ensures that the correct sentence will be read out loud in a real communication scenario.

	\section{Summary and conclusions}

	In this paper, we proposed a lip-to-speech system built from a backend for deep neural network training and inference and a fronted as a form of a mobile application.
Compared to earlier lip-to-speech and lip reading systems~\cite{Akbari2018,Ephrat2017,LeCornu2015,Michelsanti2020,Racz2020,Sun2018,Wand2016,Wand2020}, the main difference is that here we focus on the practical implementation of the whole system, and not only the deep learning aspects. A limitation of the current system is the speed, i.e.\ inference at the server and network delay make real-time communication somewhat inconvenient.
Our initial evaluation shows that although the recognition results are relatively low, the scenario is feasible because of the user-in-the-loop steps, and potential end-users might use the application. Silent Speech Interfaces are targeting the speaking impaired, e.g.\ those after laryngectomy~\cite{Denby2010,Gonzalez-Lopez2020}. Besides, automatic lipreading can be useful if we consider privacy concerns: some people do not feel comfortable if they have to speak loud to their smartphones when others are around.

	In future work, we plan to test the system with multiple users from the target user group. Other, more complex networks are also intended to be used, taking into account the fast inference and response speeds required for real-time communication.

	\section{Acknowledgements}

	The authors were partially funded by the National Research, Development and Innovation Office of Hungary (FK 124584 and PD 127915 grants).

	\bibliographystyle{splncs03}

	\bibliography{aicv2021_vid2speech_v1}

\end{document}